\documentclass{article}

\usepackage{arxiv}

\usepackage[utf8]{inputenc} 
\usepackage[T1]{fontenc}    
\usepackage{hyperref}       
\usepackage{url}            
\usepackage{booktabs}       
\usepackage{amsfonts}       
\usepackage{nicefrac}       
\usepackage{microtype}      
\usepackage{graphicx}
\usepackage{listings}

\title{Pain Intensity Estimation from Mobile Video Using 2D and 3D Facial Keypoints}

\author{Matthew Lee \\
    \textbf{Lyndon Kennedy} \\ 
    \textbf{Andreas Girgensohn} \\ 
    \textbf{Lynn Wilcox} \\
    FX Palo Alto Laboratory \\
    Palo Alto, CA, USA \\
    \texttt{matthew.l.lee@acm.org} \\
    \texttt{lyndonk@acm.org} \\
    \texttt{andreasg@acm.org} \\
    \texttt{lynndwil@gmail.com}
\And
    John Song En Lee\textsuperscript{1} \\ 
    \textbf{Chin Wen Tan\textsuperscript{1,2}} \\ 
    \textbf{Ban Leong Sng\textsuperscript{1,2}} \\
    \textsuperscript{1}Women's Anaesthesia \\
    KK Women's and Children's Hospital \\
    \textsuperscript{2}Duke-NUS Medical School \\
    \texttt{john.lee.s.e@singhealth.com.sg} \\
    \texttt{Tan.Chin.Wen@kkh.com.sg} \\
    \texttt{sng.ban.leong@singhealth.com.sg}
}

\hypersetup{
pdftitle={Pain Intensity Estimation from Mobile Video Using 2D and 3D Facial Keypoints},
pdfsubject={},
pdfauthor={Matthew Lee, Lyndon Kennedy, Andreas Girgensohn, Lynn Wilcox, John Song En Lee, Chin Wen Tan, Ban Leong Sng},
pdfkeywords={pain assessment, facial keypoints, 3D face mesh, multiple instance learning, mobile computing},
}

\begin{document}
\maketitle

\begin{abstract}
Managing post-surgical pain is critical for successful surgical outcomes. One
of the challenges of pain management is accurately assessing the pain level of
patients. Self-reported numeric pain ratings are limited because they are
subjective, can be affected by mood, and can influence the patient's perception
of pain when making comparisons.  In this paper, we introduce an approach that
analyzes 2D and 3D facial keypoints of post-surgical patients to estimate their
pain intensity level. Our approach leverages the previously unexplored
capabilities of a smartphone to capture a dense 3D representation of a person's
face as input for pain intensity level estimation.  Our contributions are a
data collection study with post-surgical patients to collect ground-truth
labeled sequences of 2D and 3D facial keypoints for developing a pain
estimation algorithm, a pain estimation model that uses multiple instance
learning to overcome inherent limitations in facial keypoint sequences, and the
preliminary results of the pain estimation model using 2D and 3D features with comparisons of alternate approaches.
\end{abstract}


\section{Introduction}
For the more than 300 million surgeries performed worldwide every year, managing post-surgical pain is critical for successful surgical outcomes. Pain is the most prominent post-surgical concern, with an estimated 86\% of surgical patients in the United States experiencing pain after surgery, with 75\% of these patients reporting at least moderate to extreme pain \cite{gan_incidence_2013}. Higher postoperative pain is associated with more postoperative complications \cite{van_boekel_relationship_2019}, indicating the importance of pain management. Furthermore, the use of opioid analgesics is a powerful tool for managing pain but can pose risks of adverse drug events (experienced by 10\% of surgical patients), leading to prolonged length of stay, high hospitalization costs, and potentially addiction \cite{urman_burden_2020}. Thus, regular and careful pain assessment is important for balancing between pain relief and potential side effects of powerful opioid analgesics \cite{skrobik_pain_2020}.

However, one of the challenges of pain management is accurately assessing the pain level of patients. Pain is inherently subjective, and the personal experience of pain is difficult to observe and measure objectively by those not experiencing it \cite{wideman_multimodal_2019} (e.g., care providers). The standard practice used in clinical care requires patients to self-report their pain intensity level using a numeric or visual scale, such as the popular Numerical Pain Rating Scale \cite{gerbershagen_determination_2011}. Though commonly used as standard practice, self-reported numeric pain ratings are limited because they are still subjective, can be affected by mood \cite{taenzer_influence_1986}, and can influence the patient's perception of pain when making comparisons \cite{lundeberg_reliability_2001}. Therefore, there is a need for more objective and unobtrusive ways of estimating pain level.

Facial expressions can be a window into people's inner subjective emotional state, including pain. In the 1970s, Ekman \& Friesen \cite{ekman_facial_1978} developed the Facial Action Coding System (FACS) to catalog how different parts of the face work together to express emotion. Prkachin \& Solomon identified the facial action units that were correlated with pain and developed the Prkachin and Solomon Pain Intensity (PSPI) metric which mapped activations of facial regions to a numeric pain score used in research and clinical practice \cite{prkachin_structure_2008}. However, manually coding facial expressions is too time consuming for clinical practice, needing approximately 10 hours of coding time per minute of behavior \cite{scherer_handbook_1982}. 

A number of automated approaches for estimating pain ratings from FACS have been developed \cite{chen_automated_2018}, but their accuracy and performance are not yet adequate for clinical use. In this paper, we introduce an approach that analyzes 2D and 3D facial keypoints of post-surgical patients to estimate their pain intensity level. Our approach leverages the previously unexplored capabilities of a smartphone to capture a dense 3D representation of a person's face as input for pain intensity level estimation. 

In this paper, we make the following contributions: 
1) a data collection study (method, apparatuses, software) with post-surgical patients to collect a dataset of ground-truth labeled sequences of 2D and 3D facial keypoints for developing a pain estimation algorithm,
2) a pain estimation model that uses multiple instance learning to overcome inherent limitations in facial keypoint sequences,
3) the preliminary results of the pain estimation model using 2D and 3D features with comparisons to alternate approaches.
\section{Background}

Pain can be characterized by intensity, onset/pattern, location, quality, aggravating factors, and functional effects. For the scope of our work, we focus on the intensity characteristic of acute transient pain because it is commonly assessed in perioperative settings. In this section, we review related work in how pain intensity level is current assessed in clinical settings, pain datasets for developing pain estimation algorithms, and prior examples of pain estimation approaches. 

\subsection{Self-Report Scales for Pain Intensity}
Guidelines for postoperative pain management recommend using validated pain assessment tools for ongoing reassessments of pain levels to track the effectiveness of pain relief regimens \cite{chou_management_2016}. Patient self-report is considered the primary basis of pain assessments because of the subjective and personal experience of pain. To report pain intensity level, a number of validated instruments for patient self-report are commonly used. The Numeric Rating Scale is typically verbally administered to patients by asking them to rate their pain intensity on a discrete numeric scale, for example from 0 to 10, with 0 corresponding "no pain" and 10 corresponding to "worst possible pain" or "worst pain imaginable." The Numeric Rating Scale (NRS) can have different length scales including 0 to 5, 0 to 10, 0 to 20, to provide a tradeoff between cognitive effort and measurement resolution \cite{morrison_pain_1998}\cite{bergh_application_2000}\cite{chibnall_pain_2001}. The NRS was found to have higher compliance rates and ease of use compared with other scales \cite{hjermstad_studies_2011}. 

The Visual Analog Scale (VAS) \cite{jensen_interpretation_2003} is a continuous variant of the Numeric Rating Scale, that asks patients to rate their pain intensity by drawing a mark on a 100mm line segment, anchored at the ends with "no pain" and "worst pain imaginable." The distance from the left edge is the pain rating. Studies have found cutoffs for interpretation in postoperative pain to be 0 to 4mm as no pain, 5 to 44mm as mild pain, 45 to 74mm as moderate pain, and 75 to 100mm as severe pain. The minimally clinically important difference was found to be approximately 10mm, with a decrease of 33 or more representing acceptable pain control after surgery \cite{myles_measuring_2017}. A digital version of the VAS on a smart tablet was found to be as reliable as the paper version \cite{escalona-marfil_validation_2020}. 

The Verbal Rating Scale (VRS) \cite{gagliese_measurement_2005} asks patients to rate their current pain intensity, as well as best and worst pain intensity over a previous time period (e.g., 24 hours) on an ordinal scale with items labeled no pain, mild pain, moderate pain, and severe pain. Even though they provide less granularity, the small number of simple response items of the VRS can be useful for those with cognitive limitations that make graphical or numeric responses difficult . The Wong-Baker Faces Pain Rating Scale \cite{wong_pain_1998} was designed initially for children who choose one of six faces ranging from a happy face at 0 to a crying face at 10 . Even though the above-mentioned self-report scales help provide structure and standardization for pain intensity levels, they are still require the patient to self-report.

\subsection{Pain and Facial Action Unit Coding}
Observer-based, behavioral methods for assessing an individual's pain intensity offer an alternative to self-report. Clinicians and informal caregivers often use their intuitive sense to estimate the pain experienced by a patient, but studies have shown observer-based ratings can be inconsistent and tend to overestimate pain levels \cite{tait_judging_2011}. Ekman \& Friesen \cite{ekman_facial_1978} developed the Facial Action Unit Coding system (FACS) for describing and measuring facial movements associated with emotions and pain. FACS decomposes facial expressions into 52 action units (AUs). Based on FACS, the Prkachin Solomon Pain intensity (PSPI) metric \cite{prkachin_structure_2008} found that a particular subset of action units (AU4 brow lowering, AU6/7 orbital tightening, AU9/10 levator contraction, AU43 eye closure) tended to be activated when people experience pain. A trained rater can look at a static image of a face (or a frame in a video) can rate the how activated each action unit is from 1 to 5 and sum up the activations for a composite pain intensity score. However, manually coding frames of video requires extensive training and is too time consuming for clinical practice. A study showed that minimally trained human raters were able to distinguish genuine from fake pain at rates no greater than chance \cite{bartlett_automatic_2014}.

\subsection{Automatic Coding of FACS and Pain Intensity Estimation}
Automatic techniques for coding FACS were developed using computer vision using both static images and across multiple frames of video. See Chen et al. \cite{chen_automated_2018} for an overview of automatic FACS coding approaches. Pain intensity estimation techniques have been developed to leverage automatic FACS coding capabilities. For example, Lucey et al. \cite{lucey_automatically_2011} first calculated frame-level pain measures by fusing the output of AU detectors using linear logical regression. Then to classify video sequences (as is done in clinical practice) as no, low or high pain, they built one-vs-all binary SVM classifiers for each pain level category, and a majority vote across frames inferred pain level was used to derive an estimate of the pain level of the sequence. Their technique was able to identify no pain, but performance for discriminating low from high pain was lacking. 

Another example, Sikka et al. \cite{sikka_automated_2015}, used a toolkit to score 10 AUs related to pain, as well as a smile AU, and head movements for each frame of video of the faces of 50 children experiencing induced transient pain (pressing a surgical site) after a laparoscopic appendectomy. Then using these features with self-reported NRS as ground truth, they trained linear regression models for binary classification for pain/no-pain (AUC 0.91) and pain intensity estimation that had moderate correlation (r=0.47) with self-reported scores. Barlett et al. \cite{bartlett_automatic_2014} applied automatic FACS coding and a SVM machine learning model to classify whether a facial expression was due to real pain or deception, with about an 85\% accuracy. They found that differences in the temporal dynamics of facial movements between genuine and faked pain.

Automatic FACS coding is used in one commercial product, PainChek, a mobile app that can estimate a pain score based on automatically analyzing the pain AUs from mobile video and other manually entered patient metadata \cite{atee_pain_2017}. PainChek extracts binary activations for 9 AUs (AU4, 6, 7, 9, 10, 12, 20, 25, 43) from a 10-second video using computer vision. The app also allows the care provider to easily input an additional 33 binary descriptors about the patient's condition including voice, movement, behavior, activity, body. A pain score is derived from summing these 42 binary inputs, with bands for no pain (0-6), mild (7-11), moderate (12-15), and severe (16-42). In a study with 353 paired assessments across 40 residents of aged care homes in Perth using both PainChek and the Abbey Pain Scale (as ground truth), they found their system had a good concurrent validity with r=0.882. A follow up study with geriatric residents showed similar results \cite{atee_psychometric_2017}. Despite the good match between human raters and the system based on automatic FACS and human input, the study did not decompose the results to a level to know how well a fully automatic face-only approach would work.

\subsection{Pain Intensity Estimation using Deep Learning}
Recent work has also explored using neural nets for estimating pain intensity from video. Liu et al. \cite{liu_deepfacelift:_2017} developed a two-stage hierarchical learning algorithm called DeepFaceLIFT, with the first stage taking facial keypoints (and other personalized features such as complexion, age, gender) as input to a fully connected 4-layer neural net with ReLu activation functions and outputs frame-level estimates of VAS pain intensity. This first stage is trained with multi-task learning with labels for the VAS pain score and an observer pain rating. The second stage takes in a sequence of estimated VAS scores (one for each frame) from the first stag, calculates statistics (min, max, median, variance, etc) across the sequence, uses a Gaussian Process Model with a RBF-ARD kernel to estimate a pain intensity (VAS) estimate for the entire sequence (video clip). The algorithm was evaluated on the UNBC-McMaster Shoulder Pain Expression Archive \cite{lucey_painful_2011} (described in section \ref{datasets}) and had a lower mean absolute error (2.18) when compared with other approaches for modeling sequential data. 

Xu et al. \cite{xu_pain_2020} used a slightly more complex multi-stage approach, relying on frame-level PSPI annotations provided in the archive \cite{lucey_painful_2011}. The approach first uses a VGGFace neural network trained to predict frame-level PSPI scores, then a fully connected neural network for multi-task learning for VAS, self-reported sensory intensity, self-reported affective-motivation, and observer rated pain, and finally an ensemble learning approach to compute the optimal linear combination of task outputs to estimate VAS. This more complex model that utilizes more labels from the dataset yielded a lower mean absolute error (1.95) compared to the prior work. For a review of other approaches, see Werner et al. \cite{werner_automatic_2019}. Despite the incremental improvements of applying new techniques to pain data, performance has yet to be adequate and reliable enough for clinical use.

\subsection{Pain Video Datasets} \label{datasets}
To develop algorithms for pain intensity estimation, labeled datasets are necessary for providing examples to learn from. One of the first publicly-available pain datasets with good annotations is the UNBC-McMaster Shoulder Pain Expression Archive \cite{lucey_painful_2011}. This archive includes videos of people's spontaneous facial expressions when experiencing various intensities of pain while performing range of motion exercises with their shoulders. The archive comprises 200 brief videos (usually <10 sec) and 48,398 frames across 129 participants. Each video (image sequence) is annotated with a self-reported VAS, self-reported sensory intensity, self-reported affective-motivation, and observer rated pain score. Each frame of video was coded using FACS by  trained experts to identify the activation of 11 action units related to pain. From these action units, the PSPI is calculated for each frame and included in the dataset. Included with each frame are also 66 facial landmarks extracted from the image using an Active Appearance Model. 

The BioVid Heat Pain Database \cite{werner_towards_2013} is another publicly available pain dataset collected from healthy participants who were stimulated with heat to induce pain at four different intensities. The BioVid database comprises 8,700 facial videos (5.5 seconds in length) for 87 participants who were stimulated 20 times at each of five intensity levels in random order. The apparatus was first calibrated for each participant's heat perception and pain threshold. The database includes biomedical signals (GSR, ECG, trapezius EMG) for each sequence. Using this database, Werner et al. \cite{werner_automatic_2017} develop a pain estimation technique that extracted facial AU activations over time from video frames, reduced the dimension of frame-level features using Principle Component Analysis, and used a Random Forest classifier to predict the pain intensity level for the video. The resulting performance is better than chance but has good potential to be improved.

The multimodal EmoPain dataset \cite{egede_emopain_2020} is a recently released dataset that includes facial video, motion (joint angles and angular velocities), and muscle activity from 18 people with chronic lower back pain and 22 healthy people. Each participant performed a series of physical activities such as sit-to-stand, stand-to-sit, reach forward, etc with and without holding 2kg weights. A baseline model was developed using OpenFace facial landmarks, head pose, FACS activations, and emotion-related features from deep learning, and resulted in a mean absolute error of 0.91 on the test set. 

Existing datasets are limited in their scope and extent of clinical applicability. No datasets are available for post-surgical pain which typically involves changes in pain/functioning as well as the use of anesthesia. Furthermore, existing datasets are limited to simple video as the primary media. In our work, we aim to explore the specific context of post-surgical pain using new modalities such as 3D facial feature detection found on new smartphones. With these new data, our aim is to develop objective pain intensity estimation techniques that are reliable and tailored for clinical use.

\section{Method}
To develop a robust pain intensity estimation technique for post-surgical patients, we designed and carried out a data collection study to generate a dataset of different types of facial keypoints and metadata of post-surgical patients experiencing acute transient pain. We describe the data collection in this section. We use this dataset to develop models for estimating pain intensity using machine learning (described in section \ref{modeling}). We investigated two approaches: the first using a two-stage hierarchical model similar to that of Liu et al. \cite{liu_deepfacelift:_2017} and the second using Multi-instance Learning (MIL) to overcome the sparseness of facial expressiveness in pain videos. 

\subsection{Clinical Setting and Participants}
Our work focuses on acute transient pain experienced in post-surgical settings and assisting clinicians manage post-surgical pain and anesthesia. Our data collection was carried out in the inpatient setting at the Department of Women's Anesthesia at KK Women's and Children's Hospital in Singapore. The data collection study and algorithm development research received human subjects ethics approval from the SingHealth Centralized Institutional Review Board (Ref No. 2019/2293) and registered in Clinicaltrial.gov with identifier NCT04011189.

The inclusion criteria for patients were: 1) undergoing major gynecological surgery, 2) expected to be prescribed morphine patient controlled analgesia post-operatively, and 3) have an American Society of Anesthesiologists physical status of 1 or 2. Exclusion criteria were: 1) currently pregnant, 2) expected to be discharged in fewer than 48 hours after surgery, 3) expected to be administered neuraxial anesthesia during surgery or not on morphine controlled analgesia post-operatively, and 4) having medical problems or use of medications including psychiatric disorders, neurological disorders, musculoskeletal limitations that result in gait abnormalities/limitations, presence of chronic pain (>3 months), and on long-term pain medications (>3 months). All patients were female and limited to an age range of 21 to 70 years old. Study participants were sourced via referral by an attending healthcare professional and were recruited by clinical research coordinators who approached patients in preoperative clinics and wards using study brochures and answered questions. Participation in the study was completely voluntary and did not change the patient's treatment plan.

In total, 27 patients were recruited into the study, with one deciding to withdraw from the study after surgery, leaving 26 who completed at least one pain assessment in each of three timepoints (pre-surgical and two post-surgical). See the following section for details on the data collection procedure. The average age of participants was 47 years old (SD=11). Participants' race backgrounds were 14 Chinese, 7 Malay, 3 Indian, and 2 Filipina.

\subsection{Data Collection and Pre-processing}
We developed a data collection protocol to record video of the patient's face, 3D facial keypoints from the mobile phone, and other patient metadata such as the self-reported pain intensity score. Post-surgical pain is often assessed both at rest and during movement \cite{srikandarajah_systematic_2011}, so we selected a combination of stationary and moving actions. In our protocol, patients were asked to verbally rate their maximum pain intensity level using the numerical rating scale from 0 (no pain) to 10 (worst pain ever) when performing five different actions in the following order: 1) seated at rest, 2) deep breath (taking a deep breath and holding for a count of 3 before exhaling), 3) sit-to-stand, 4) standing, and 5) stand-to-sit. Patients were asked to perform these ratings at three different time points or sessions: before surgery, 12-36 hours after surgery, and 36+ hours after surgery (before discharge). At any time, patients could decline to perform an action (and associated pain rating), and the clinical research coordinator would skip that action. A patient who performed all five actions in each of the three sessions would generate 15 data samples labeled with their self-reported pain score.

\subsubsection{Data Collection with a Mobile Phone}
\begin{figure}[h]
    \centering
	\begin{tabular}{@{}c@{\hspace{0.5in}}c@{}}
        \frame{\includegraphics[width=1.65in]{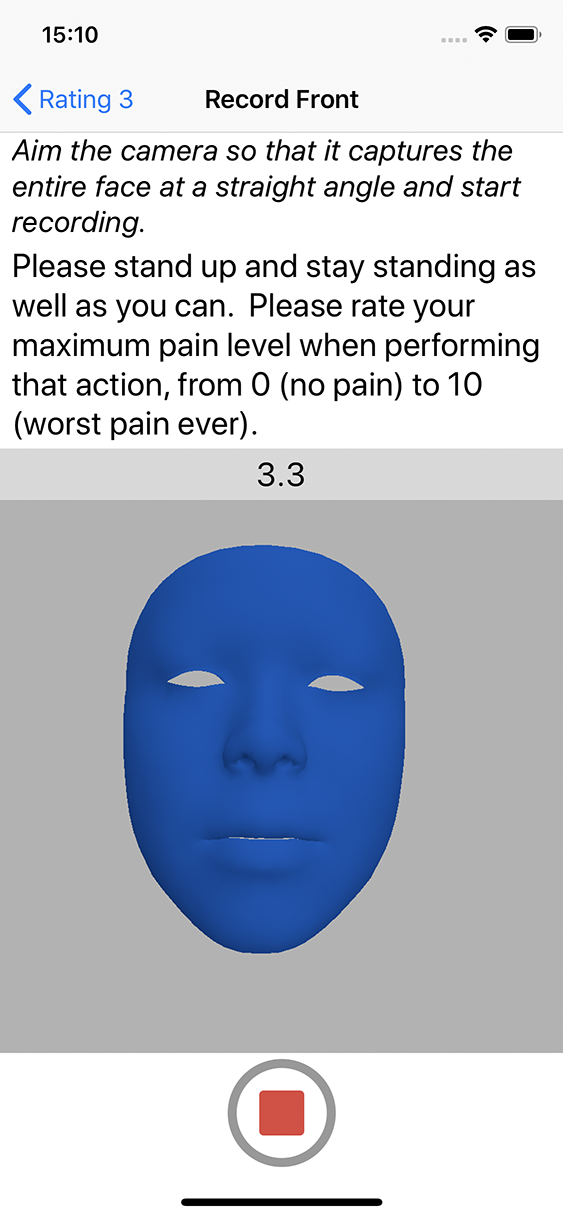}} &
        \frame{\includegraphics[width=1.65in]{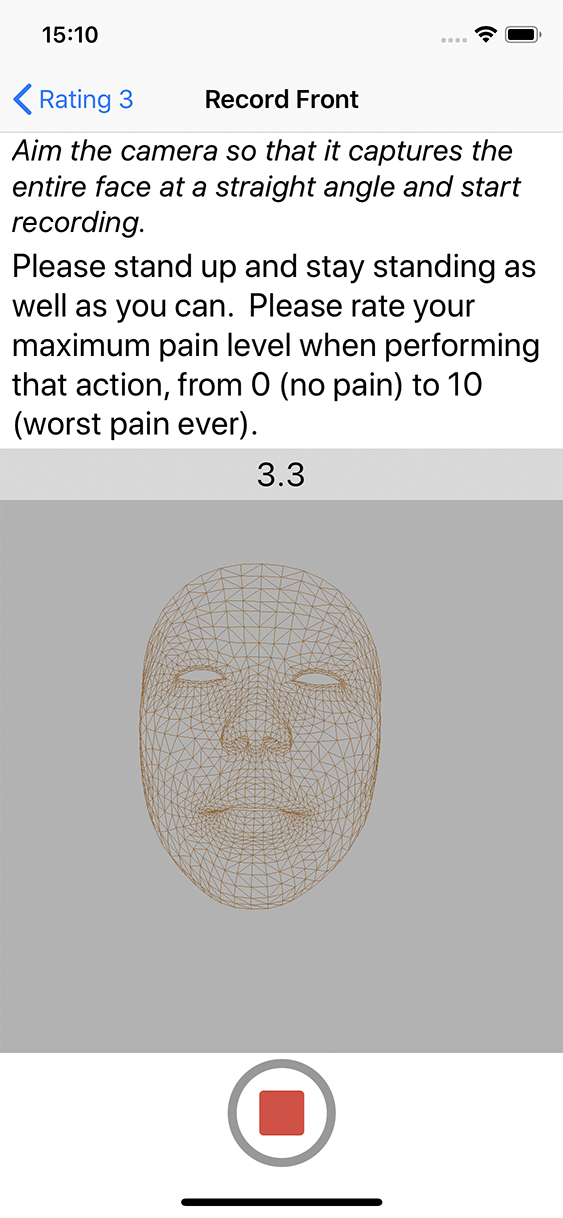}}
    \end{tabular}
	\caption{Custom iPhone app used for collecting pain video data. Standard prompts are shown for the research coordinator to say aloud to the patient, while a live representation of the patient's face is shown as feedback to keep the face in the camera frame. Solid and wire-frame face representations are available.}
	\label{fig:capture-app}
\end{figure}

To assist research coordinator to record data consistently, we developed a custom mobile app (Figure \ref{fig:capture-app}) for the Apple iPhone. The app presented a dialog tree for each patient that structured the data collection according to the protocol and automatically recorded multiple streams of data necessary for algorithm development. For privacy purposes, no patient identifiers were stored on the iPhone, with the data indexed only by a participant ID assigned only for this study. The app allowed the coordinator to select the study participant ID, session, and action to be performed by the patient. For each action, the app would display the verbal prompts for the coordinator to instruct the patient and ask for the pain score. 

During each action, the app recorded video of the patient's face while the research coordinator wrote down the reported pain score (and later entered it into the research database). The video of the patient's face was recorded with an iPhone XR (running iOS version 12.3.1). Most video was recorded with the iPhone's TrueDepth camera consisting of a structured light transmitter and receiver, a front-facing camera, and a time-of-flight proximity sensor. In addition to the video, the camera provided a 3D mesh of the face and BlendShape coefficients via Apple's ARKit API that we recorded as well. To evaluate an alternate capture method, video for the deep breath action was recorded with iPhone's rear camera that does not provide a 3D face mesh.


\begin{figure}
    \centering
	\includegraphics[width=2.5in]{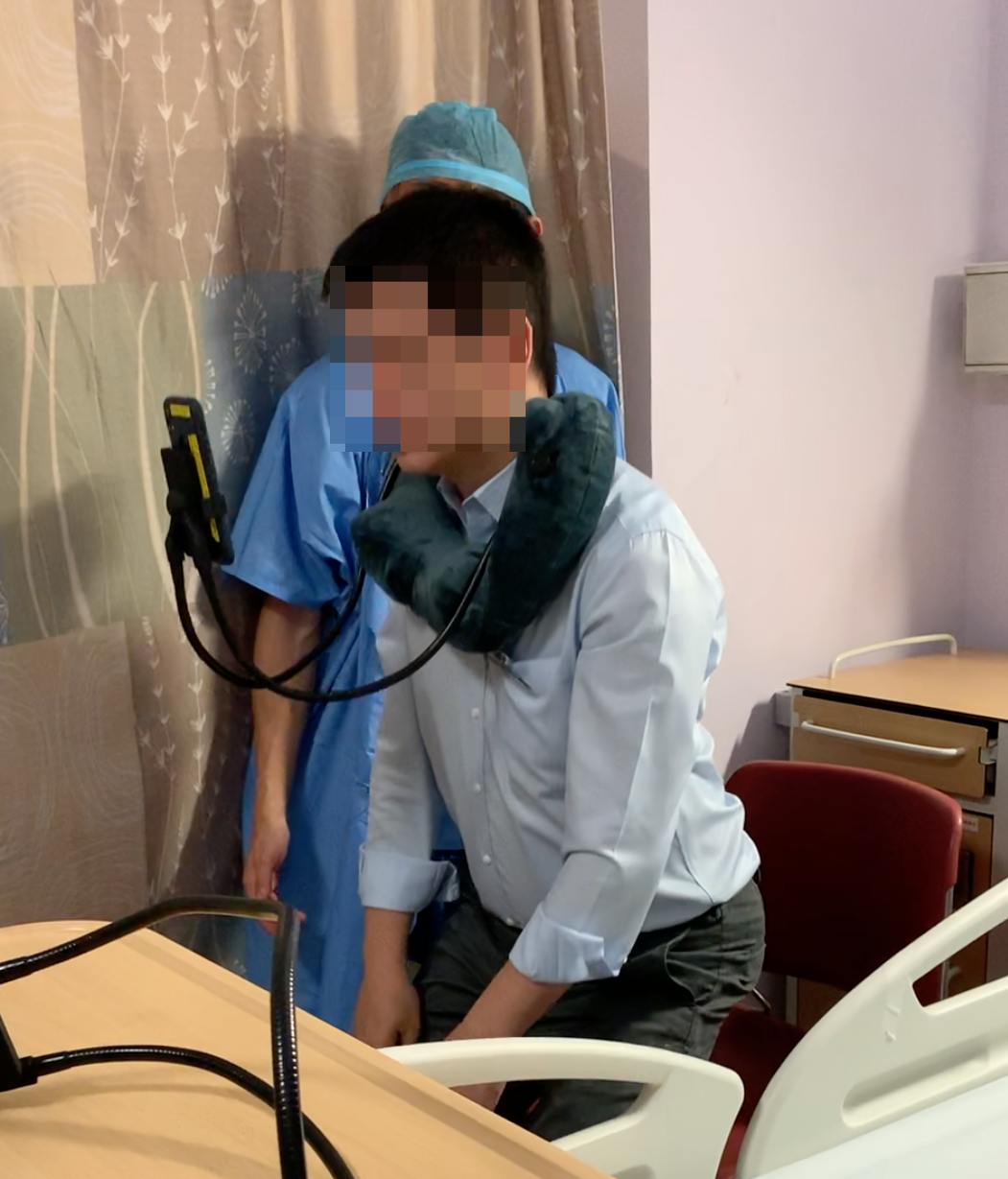}
	\caption{To stabilize the image and keep the patient's face in the frame, the iPhone was mounted on a holder that extended from the neck and was cushioned by a U-shaped neck pillow above the patient's sternum.}
	\label{fig:phoneholder}
\end{figure}

For the first action, seated at rest, patients held the phone pointing the phone at their face while verbally answering on-screen survey questions, as the phone recorded selfie video with the front camera. For the deep breath action, the research coordinator recorded the patient with the rear camera. The iPhone's TrueDepth camera is optimized for use within arm's length and was difficult for the research coordinator to hold steady and track the patient's face. Thus, for the remaining three actions (sit-to-stand, standing, stand-to-sit), we used a phone holder worn around the neck and cushioned by a U-shape neck pillow resting on patient's sternum (Figure~\ref{fig:phoneholder}). The holder positioned the phone roughly 50~cm in front of the patient's face and kept the face in view of the TrueDepth camera while they performed the actions. 

Providing live feedback of the video or face helped patients keep their face within the video frame. To overcome people's tendency to smile when seeing their own face, we displayed a live abstraction of the 3D face mesh on the iPhone's display. Figure~\ref{fig:capture-app} shows two of the abstractions that could be chosen in the settings of our app, a solid, reflective texture, and a wire frame. For the data collection study, we selected the wire frame because it looked the most natural and least distracting to the patients.

Video was recorded in high definition (1080x1920 back; 1080x1440 front) in YUV420p at 60 fps. For the face mesh data, we recorded data provided by ARKit 
ARFaceGeometry instances.  Those represent several transformation matrices and the vertices, triangles, and textures of 1,220 3D face keypoints. As the video, those instances arrive at 60 fps. A first attempt to record the face mesh data as JSON with floating-point values for each vertex had very poor performance. As an alternative, the array of floating-point values was copied as binary data and encoded as Base64 in the JSON file. Keeping a whole recording session in memory and storing it at the end was not possible for longer sessions. Instead, face mesh data was stored as a separate JSON file every ten seconds in a background thread. Those files were combined in a Zip file for transfer and merged for processing by the machine learning component.

\subsubsection{Pre-processing the Collected Data}
For privacy reasons, the iPhone was not connected to the network, and the recorded video was transferred from the iPhone to an external hard drive encrypted with a passcode via a non-networked PC located at the hospital. Transfer from the iPhone and processing of the data was mostly automated with several shell scripts. Data transfer was facilitated by mounting the capture app portion of the iPhone file system with the \texttt{ifuse} command and by synchronizing the external drive with that file system.

To comply with hospital guidelines, video data was not allowed to leave the hospital premises. The recorded 3D face data was deemed to be sufficiently de-personalized to be used for selective research outside the hospital. To investigate alternative approaches for situations where 3D face data is not available, we processed the video to extract sufficiently de-personalized facial keypoints.

We chose OpenPose~\cite{cao_openpose_2019} that can detect 70 facial keypoints. We created a pipeline to process the video on the non-networked PC equipped with an NVIDIA GeForce RTX 2080 Ti graphics card. Facial keypoints were detected at a resolution of 304x304 at five different scales, the most that fit into 11 GB graphics memory. Higher resolutions did not improve the results but a larger number of scales did. OpenPose provides a 70-keypoint facial keypoint estimation per frame. The TrueDepth camera provides a 3D mesh with 1,220 vertices at 60 fps. Both approaches use consistent positions for keypoints in the face such as the tip of the nose or the corners of the eyes. By having both OpenPose facial keypoints and a 3D mesh for the same video sequence, we can directly compare the performance of both data sources.

\begin{figure}[t]
    \centering
	\includegraphics[width=3in]{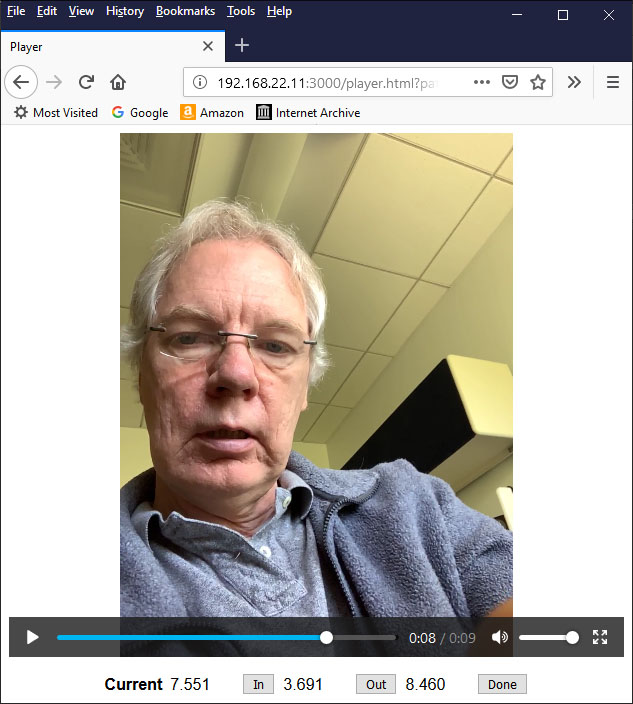}
	\caption{UI for marking start and end in a video clip so irrelevant data could be trimmed later during data cleaning and modeling.}
	\label{fig:video-clipper}
\end{figure}

To make sure that only the actual sessions were included in the data, a web-based UI running on the PC may be used by authorized staff to clip the video by marking the start and end (see Figure~\ref{fig:video-clipper}). This UI consists of a video player and buttons to mark an endpoint at the current video position. The machine learning component ignores data outside those positions.

The facial keypoint output from OpenPose, the 3D face mesh, and the video clipping data was copied as a Zip files encrypted with GNU Privacy Guard (GPG) onto a USB drive. Data was identified only by a participant ID assigned only for this study. No other patient identifiers (e.g., patient name) were included in the dataset. To transfer the data, the USB drive was carried from the hospital to an office of the researchers' employer and securely stored in password-protected local storage. This provided privacy protection by several means, de-personalized facial keypoints, encryption, and avoidance of publicly accessible networks.


\subsection{Summary of Dataset}
The data collection study resulted in a dataset with a total of 319 pain ratings from 26 gynecological surgery patients and 3 pain assessment time points. For each pain rating, the dataset contains the following:
\begin{itemize}
   \item 70 facial keypoints extracted from each frame of facial video using OpenPose,
   \item 3D face mesh (1,220 facial keypoints in 3D for each frame) recorded in real time with the iPhone XR TrueDepth camera using Apple's ARKit API
   \item BlendShape coefficients (corresponding to 52 different facial actions) for each frame using Apple's ARKit API
   \item a patient-reported numeric pain intensity score from 0 (no pain) to 10 (worst pain ever)
\end{itemize}

For details on the data formats, see Appendix \ref{dataformats}.

For each patient, the dataset includes the following demographic information: age, race, height, weight, and medical procedure.

For each assessment time point (preoperative, postoperative 12-24 hours, and postoperative 36 hours to discharge), the dataset includes the following patient questionnaire data: anxiety and depression scores from the Hospital Anxiety and Depression Scale~\cite{bjelland_validity_2002} and ratings for mobility, self-care, usual activities, pain/discomfort and anxiety/depression from responses to the EQ-5D-3L~\cite{szende_self-reported_2014}.

At publication date, this dataset has not been made publicly available. 

\section{Face Representation and Pain Modeling} \label{modeling}

To demonstrate the efficacy of the data and features collected in our dataset, we apply several baseline implementations of pain prediction algorithms on various feature subsets. Specifically, we directly apply the DeepFaceLIFT model from Liu et al. \cite{liu_deepfacelift:_2017}. and expand upon it with some further refinements. We also apply a version of Multiple Instance Learning (MIL) to address the nature of the data and task more directly. In each of these cases, we also explore estimating pain levels directly through regression models or changing to a binary prediction on significant or insignificant pain levels. We demonstrate that the 3-dimensional features available in this dataset show promise in terms of performance relative to 2-dimensional, video-only features. We further find that approaching the problem directly with Multiple Instance Learning shows benefits in this case where the available data is small and the task is limited to binary prediction.

\subsection{Data Representation}

The samples in our dataset are represented as sequences of face images. The sequences tend to be 5-10 seconds in length and each has a final sequence-level numeric label of the pain experienced by the subject ranging from 0 to 10. The data are prepared for use in the various machine learning methods as follows.

\subsubsection{Input Features}

\paragraph{2D Keypoints.}

OpenPose provides 70 2-dimensional points corresponding to facial landmarks in each frame. The values of these x,y coordinates map to the pixels in the image. We normalize these points to fall between 0 and 1 by calculating the maximum bounding box containing all face points on each frame and scaling the points relative to their distance from the edges of the box. OpenPose further yields a confidence value for each predicted point, scaled between 0 and 1. We combine the 70 normalized 2-dimensional points along with the confidence score to yield a 210-dimensional feature vector for each frame in the dataset. In the rare cases where there are multiple faces detected in a given frame, we default to use the points corresponding only to the largest detected face.

\paragraph{3D Keypoints.} 

The TrueDepth ARKit API provides 1,220 3-dimensional points per frame. The coordinates of the points arrive already normalized between -1 and 1, which map to the Euclidean distance of the point from the center of the face. The API provides further information for scaling and rotating the face within the viewport of the camera, but these normalized representations are already suitable for learning, so we ignore this information. The 1,220 3-dimensional points are then flattened into a 3660-dimensional feature vector for each frame in the dataset. Again, if there are multiple faces in the frame, the API will only return the most prominent.

\paragraph{BlendShapes.} 

The TrueDepth ARKit API also provides 52 scalar values, each representing the extent of various facial actions, such as mouth opening, eye closing, or brow lifting. The values already arrive normalized between 0 and 1 and are used directly as a 52-dimensional feature vector for learning purposes.

\subsubsection{Output Targets}

\paragraph{Direct Pain Level Prediction.} 

In some sets of experiments, we learn to predict the pain level of the sequence directly by regression methods. The raw pain levels collected are integers between 0 and 10. We scale them to floating-point numbers between 0 and 1 by dividing by 10.

\paragraph{Binary Prediction.}

In some sets of experiments, we predict  ``significant’’ and ``insignificant’’ pain by thresholding the raw pain score for each sequence at a pain level of 4 \cite{gerbershagen_determination_2011}.

\subsection{Deep Network Approaches}

To test the DeepFaceLIFT model, we first implement the model as reported by Liu et al. \cite{liu_deepfacelift:_2017}, using identical network size and experimental settings in terms of batch size, number of epochs, etc. We replicate the reported experiments on the same shoulder pain dataset \cite{lucey_painful_2011} and confirm similar performance on the same task. 

After we have confirmed the efficacy of the DeepFaceLIFT model on the shoulder pain dataset, we turn towards also testing it on our data. The shoulder pain dataset contains facial 66 2-dimensional keypoints from an Active Appearance Model. This is similar, though not entirely equivalent to the 70 2-dimensional keypoints that we derive from running OpenPose over our dataset. The points modeled are similar in their location, but OpenPose returns an extra 4 landmarks and we also incorporate the confidence value.

\subsubsection{First Level}

In the first level of the DeepFaceLIFT model, the training samples are randomly chosen frames from the training set. The input to the model is the vector representation of the features and the target to learn is the sequence-level pain score. The labels of the sequence are applied equally to each frame of any given sequence, which we can interpret as a form of weak labeling. The model consists of four fully-connected layers and uses a mean-squared error loss function to estimate the pain score in the final output. We experimented with inserting two dropout layers, with a dropout rate of 50\%, after the second and third layers, which showed some minor improvement in training stability. The results reported here include this modification.

\subsubsection{Second Level}

At test time, we can feed through a new unseen frame through the trained network and arrive at a predicted pain score for each frame. Test sequences, however, consist of hundreds of frames, so we require the second level of the DeepFaceLIFT model to aggregate from these hundreds of frame-level predictions to a single sequence-level prediction. A simple baseline is to simply take the maximum predicted pain score from the sequence as the overall predicted pain level. The DeepFaceLIFT model takes a number of other statistics over the predicted scores, including the mean, median, minimum, and variance to characterize the sequence and fits a Gaussian process model to predict the final sequence-level score. We further propose to augment this approach by replacing the Gaussian process with a support vector regressor to get a more sensitive and accurate model.

\subsubsection{Classification Task}

In the process of developing these models, we further investigate the applicability of a simple classification. In clinical settings, a pain score below or equal to 4 might be interpreted as an acceptable level of pain, requiring no further intervention, while a score above 4 might indicate a significant level of pain requiring further intervention \cite{gerbershagen_determination_2011}. To model this, we binarize the raw pain score into these two levels and replace the second level of the DeepFaceLIFT model with a support vector classifier (using identical aggregation features as the Gaussian process and support vector regressor models on the predictions from the first level) to predict a binary classification into either of these classes. We apply a sliding threshold across the value of the distance from the decision boundary to arrive at a receiver operating characteristic curve that can be used to tune the desired sensitivity for false alarms and missed detections for later application.

\subsubsection{Experimental Settings}

To train and test these models, we train each level using a leave-one-out approach, where all of the sequences for one patient are held out for the test set and the sequences from the remaining patients are used for the training set. We iterate through all patients until each has been individually held out in a test case. This has the advantage of allowing for a larger training set from a constrained data set. It also assures that the model is always being tested on a patient whose face the model has never seen before, which mimics the real-world application of the model in a clinical setting and assures that the model is not capturing features unique to the behaviors of individuals, but rather the common expressions of pain across many people.

\subsection{Multiple Instance Learning Approaches}

In the process of applying the DeepFaceLIFT approach and other variants, we observe some interesting phenomena about the data. Primarily, while many video sequences yield a high pain score, that high pain is only experienced by the subject for a short sub-segment of the sequence and that experience of pain is only visible in their facial expression for a few frames out of a sequence of several hundred while the remaining expressions in the sequence are relatively neutral. This means that within a 10-second clip of someone experiencing and expressing extreme pain, the majority of the frames in the clip could be indistinguishable from a completely neutral expression clip with no experienced pain at all.

Multiple Instance Learning (MIL)~\cite{dietterich1997solving} is a broad class of machine learning approaches driven by scenarios where a complete labeling of individual item instances is unavailable, but there are sets of items where it is known if at least one of the items in the set is from a particular class or not. A common metaphor for describing MIL is key rings. We can consider a dataset of key rings wherein our labels are applied at the key ring level, telling us whether each key ring has at least one key that will open a target lock. The labels do not tell us, however, which key precisely is the one that opens the lock.

In our dataset, pain expressions are seen individually at the frame level, while the labels for the level of pain are gathered for the entire frame sequence. So, if we treat each video sequence in our dataset as a ``bag'' composed of a sequences of ``instances'' of frames, then our dataset maps directly to MIL frameworks.  In fact, in some sense, the first level of the DeepFaceLIFT model might be considered as a version of MIL known as single instance learning, where we directly apply the bag score to each instance directly and aggregate individual instance scores to predict bag level scores.

There are a number of formulations and implementations of MIL. For our experiments, we choose MI-SVM~\cite{andrews2003support}, which works by relaxing the constraints of support vector machines to fit the MIL problem.  There is an open source implementation~\cite{doran2014theoretical} freely available~\footnote{https://github.com/garydoranjr/misvm}, which is helpful for reproducibility. 

\subsubsection{Sampling Strategies}
The sequences of frames that we have gathered, at 60 frames per second, are very dense. The computational cost of including all of the frames in the data while learning is prohibitive and the relative benefits of completeness are likely limited since the frames are temporal in nature and the inter-frame changes in expression and information are likely minimal. Therefore, it is necessary to sample $k$ frames from each sequence of $n$ frames in order to construct bag representations for each frame sequence.

\paragraph{Random Sampling.}
A first-order approach to sampling might be to simply extract a random set of frames from the sequence. This can be achieved by randomly permuting the collection of frames in a sequence and choosing the first $k$ frames from the resulting permutation.

\paragraph{Uniform Sampling.}
If the expression of pain is extremely momentary, then random sampling strategies might miss the relevant frames expressing that pain. An alternate approach might be to sample temporally uniformly across the sequence. This is achieved by sampling $k$ frames spaced out across the sequence with an equal number of skipped frames in between.

\paragraph{Cluster-based Sampling.}
While uniform sampling can ensure that informative positive segments are not missed, it comes at the cost of possibly oversampling uninformative negative frames across the length of the clip. To address this, we might conduct a segmentation of the video clip into portions where the facial expression within each subsegment is visually consistent. To arrive at such a segmentation, we apply a temporal agglomerative clustering. Each frame is represented by feature-space representation of the face it contains and temporally-adjacent frames are greedily joined into clusters to arrive at visually-consistent segments until the desired number of $k$ clusters is reached. The center frame from each segment is then sampled to give the final samples for the sequence.

\subsubsection{Experimental Settings}

Again with the experiments for MIL, we set binary targets by thresholding raw pain scores at 4. We perform many random train/test splits where each patient’s clips all fall either in the training or test set to avoid influence by the model having prior knowledge of any given patient’s face. We experiment with values of $k$ for sampling and find that performance saturates at $k=30$ while learning is still very fast. All results reported here use $k=30$.

\section{Results and Discussion}

For each of the pain modeling approaches that we have described, we run experiments using each of the three available feature spaces as inputs separately. We separate the experiments into two broad classes: those that estimate pain score directly and those that predict binary ``insignificant'' versus ``significant'' pain levels.

Table~\ref{tab:regressionresults} shows the results of the direct pain level regression models expressed in terms of Mean Absolute Error (MAE). Each is simply a DeepFaceLIFT with a different variant for the second level of score aggregation. We see that performance of maximum-value aggregation (``Max'' in the table) is relatively the same across all feature types. We further see an strong relative improvement using a Gaussian Process (GP) aggregation, though the effect is still consistent across all feature representations. The Support Vector Regression (SVR) approach outperforms all and also indicates some advantages to the BlendShapes and 3D Keypoints over the 2D Keypoints. MAE is directly interpretable as the average error in predicted score. So, on a scale of 0-10, the predicted values using a support vector regressor with the DeepFaceLIFT model gives predictions that deviate from the true value by 1.4 points, on average.

\begin{table}
  \centering
  \begin{tabular}{l | l  l l l}
  \toprule
  & \textbf{Max} & \textbf{GP} & \textbf{SVR} \\
  \midrule
\textbf{2D Keypoints} & 1.996 & 1.635 & \textbf{1.431} \\
\textbf{3D Keypoints} & 1.999 & 1.631 & \textbf{1.394} \\
\textbf{BlendShapes} & 1.999 & 1.633 & \textbf{1.355} \\

  \bottomrule
  \end{tabular}  \caption{Mean Absolute Error scores for variants of DeepFaceLIFT approach to directly estimating pain scores. Each implementation uses the same network for first-level prediction and then alternately uses maximum (Max), Gaussian Process (GP), or support vector regression (SVR) for second-level final prediction. }~\label{tab:regressionresults}
\end{table}

Table~\ref{tab:binaryresults} shows the results of the binary classification models. These include our variant of DeepFaceLIFT that uses an SVM classifier for the secondary level (``DFL-Binary'' in the table) alongside MIL with the various sampling approaches. We see that the MIL methods tend to perform the best. We further observe that our proposed temporal clustering approach provides benefits over other naive sampling methods.

\begin{table}
  \centering
  \begin{tabular}{l | l  l l l}
  \toprule
  & \textbf{DFL-Binary} & \textbf{MIL-Cluster} & \textbf{MIL-Random} & \textbf{MIL-Uniform} \\
  \midrule
\textbf{2D Keypoints} & 0.690 & 0.678 & 0.692 & \textbf{0.710} \\
\textbf{3D Keypoints} & 0.724 & \textbf{0.750} & 0.727 & 0.713\\
\textbf{BlendShapes} & 0.691 & \textbf{0.826} & 0.754 & 0.756 \\

  \bottomrule
  \end{tabular}  \caption{Area Under Curve (AUC) performances for each of the three feature types using a binary classifier with DeepFaceLIFT (DFL) and multiple instance learning (MIL) for each of three different sampling approaches. }~\label{tab:binaryresults}
\end{table}

\subsection{3D vs. 2D}

The 3D keypoints and the BlendShapes features are both derived from 3-dimensional range scans of the face and benefit from the depth information that is captured, while the 2D keypoints do not. Across all of the tasks, we consistently see the 3D features outperforming the 2D features. This is promising since these 3D features are a unique component of our collected dataset and it is hoped that they can provide additional capability in detecting pain in facial expressions.

\subsection{Multiple Instance Learning}

We observe a consistent benefit from formulating the pain modeling problem as a multiple instance learning task when the target is binary pain level prediction. Some possible reasons for this might be that the alternate approach relies on a roundabout way to address the same issue of many low-information frames being mixed in with a few high-information frames, while MIL attacks this directly. Similarly, the deep network used in the alternate approach might not have the capacity to learn discriminative representations of faces given the relatively small amount of training data available and a more classical approach to machine learning might be better suited to the task.

\subsection{Dimensionality}

Across the various tasks, we observe that the BlendShapes features frequently outperform the other features. One possible explanation is that these features capture facial expressions explicitly, while the other features capture facial landmarks and require the models to internally learn intermediate representations that might be useful for encoding facial expressions. Another explanation is that they are relatively compact (52 dimensions, compared to 210 or 3660), so they suffer less from the ``curse of dimensionality,'' wherein distances in increasingly large spaces become less meaningful for representing actual similarity. This effect is particularly salient when viewing the effect of using a clustering-based sampling for multiple instance learning. The clustering method works directly on this input space, so the effect is minimal on the high-dimensional features, but is very prominent on the relatively low-dimensional BlendShapes. 
These findings can suggest the need for more data to overcome the difficulty when working with high dimensional data. Another approach could be using a different source of 3D face data (even from another domain or task such detecting emotion or fraud) to initialize the model with intermediate representations re-usable for pain intensity estimation.

\section{Conclusion}
We presented some steps towards methods for automatically predicting post-surgical pain intensity using facial images captured with a mobile phone. In particular, we have described a data capture method using sensors on readily-available commercial phones to capture both standard 2-dimensional RGB video and dense 3-dimensional facial landmark representations with a repeatable and reliable apparatus. 
We have collected a pain dataset with surgical patients experiencing multiple instances of pain while performing various actions in a clinical setting where pain management is critical. 
We have further applied recent methods to this dataset to demonstrate its utility for face pain prediction. We find that these methods offer some promising early results and demonstrate the efficacy of using 3-dimensional features over 2-dimensional features. We further propose methods for predicting binary pain levels using multiple-instance learning and show improvements using these approaches. 
Our methods, dataset, and modeling lay the groundwork for future automatic pain intensity estimation techniques on easy-to-use mobile platforms that enable clinicians and informal caregivers to assess the pain level of patients and provide them optimal care.

\bibliographystyle{unsrt}
\bibliography{references}

\newpage
\appendix

\section{Data Formats} \label{dataformats}

All data is stored in JSON files.  Files have the following naming pattern: \texttt{A\_B\_C\_YYYY-MM-DD\_HH-MM-SS*.json} where \texttt{A} is the patient
number, \texttt{B} is the collection number, and \texttt{C} is the rating 
number.  ``*'' may be empty for the file describing the video or be of the 
form \texttt{\_face\_N} for face data chunks.  Files are stored in a 
directory hierarchy \texttt{pA/B/C}.

\subsection{Video Information}

For each video, there is a file with the properties \texttt{patient}, 
\texttt{collection}, \texttt{rating}, \texttt{start}, \texttt{duration},
\texttt{timestamp}, \texttt{frameTimestamps}.  The first three properties
correspond to the properties \texttt{A}, \texttt{B}, \texttt{C} above. 
\texttt{start} is an ISO date string in millisecond precision.  The 
remaining properties represent floating-point seconds. 
\texttt{frameTimestamps} is an array with a timestamp for each video 
frame. Timestamps are represented as iOS \texttt{TimeInterval}.

\begin{lstlisting}
{
  "patient": 2,
  "collection": 1,
  "rating": 4,
  "start": "2019-08-04T03:04:13.906Z",
  "duration": 16.941,
  "timestamp": 161196.78
  "frameTimestamps": [161196.78, 161196.797, 161196.813,  ...]
}
\end{lstlisting}

\subsection{Face Data}

The 3D face mesh data is stored in 10-second chunks in JSON files.  
Those files also have the properties \texttt{patient}, \texttt{collection},
\texttt{rating}, \texttt{start}.  In addition, they have the property
\texttt{blendShapeLocations}, an array of strings describing the locations 
of blend shapes.  They also have the property \texttt{data} with an array 
of face data. The face data is a direct representation of ARKit 
ARFaceGeometry instances (\url{https://developer.apple.com/documentation/arkit/arfacegeometry}).

\begin{quote}
Face mesh topology is constant across ARFaceGeometry instances. That is, 
the values of the vertexCount, textureCoordinateCount, and triangleCount 
properties never change, the triangleIndices buffer always describes the 
same arrangement of vertices, and the textureCoordinates buffer always 
maps the same vertex indices to the same texture coordinates.
\end{quote}

Each face data entry has a \texttt{timestamp} property with the same 
representation as that for video frames. Some properties such as 
\texttt{leftEyeTransform} are represented as JSON arrays of floating-point
numbers. For other properties such as \texttt{textureCoordinates}, the 
binary data of the float32 and int16 arrays is stored as Base64-encoded 
strings in the JSON data. Even though \texttt{triangleIndices} don't 
change, they are still stored in every data entry.

\begin{lstlisting}
{
  "patient": 2,
  "collection": 1,
  "rating": 4, 
  "start": "2019-08-04T03:04:13.906Z",
  "blendShapeLocations": ["browDown_L", "browDown_R", ...],
  "data": [
    {
      "timestamp": 161196.797, 
      "transform": [[0.0137, ...], ...],
      "cameraTransform": [[1, 0, 0, 0], ...],
      "leftEyeTransform": [[0.9995, ...], ...],
      "rightEyeTransform": [[0.9989, ...], ...],
      "lookAtPoint": [0.0071, ...],
      "blendShapes": "NzyY...",
      "vertices": "TZyX...",
      "textureCoordinates": "QwAA...",
      "triangleIndices": "sQQs..."
    },
    ...
  ]
}
\end{lstlisting}
The following code 
snippet shows how to extract the data with Numpy.

\begin{lstlisting}
import base64
import numpy as np

class FaceData:
    def __init__(self, data):
        for k in ['timestamp']:
            if k in data:
                setattr(self, k, data[k])
        for k in ['transform', 'cameraTransform', 'leftEyeTransform',
                  'rightEyeTransform', 'lookAtPoint']:
            if k in data:
                setattr(self, k, np.array(data[k]))
        # Note that vertices are simd_float3, each with 4 bytes padding.
        for k, t, c in [('blendShapes', np.float32, 1),
                        ('vertices', np.float32, 4),
                        ('textureCoordinates', np.float32, 2),
                        ('triangleIndices', np.int16, 3)]:
            if k in data:
                x = base64.b64decode(data[k])
                y = np.frombuffer(x, dtype=t)
                if c > 1:
                    y = np.reshape(y, (-1, c))
                setattr(self, k, y)
\end{lstlisting}

\subsection{Pose Data}

Pose data is computed by OpenPose (\url{https://github.com/CMU-Perceptual-Computing-Lab/openpose/blob/master/README.md}) and stored in its JSON format. Data directories
with pose data are stored in the \texttt{pose} directory in the same hierarchy as 
the other data. An additional directory is created for each video with a name \texttt{A\_B\_C\_YYYY-MM-DD\_HH-MM-SS}. Inside that directory are files \texttt{A\_B\_C\_YYYY-MM-DD\_HH-MM-SS\_NNNNNNNNNNNN.json}. The directory and the 
files are named by the start time of the video. Each file contains a 
property \texttt{people} with an array representing detected people. Each data 
entry has the properties \texttt{pose\_keypoints\_2d} and \texttt{face\_keypoints\_2d} 
with arrays of floating-point numbers as sequences of x-coordinate, y-coordinate, 
confidence. Those arrays may be empty. Otherwise, they contain coordinates of key 
points in a fixed order.

\begin{lstlisting}
{
  "version": 1.3,
  "people": [
    {
      "person_id": [-1],
      "pose_keypoints_2d": [
        468.369, 803.854, 0.702162,
        533.131, 1324.73, 0.409137,
        ...
      ],
      "face_keypoints_2d": [
        251.528, 736.052, 0.665484,
        261.383, 797.641, 0.715532,
        ...
      ]
    },
    ...
  ]
}
\end{lstlisting}

\subsection{Stage Directory}

The stage directory stores data ready for export to an external USB drive.  It only
has one level of subdirectories for patients.  It contains three types of files, all
with the prefix \texttt{A\_B\_C\_YYYY-MM-DD\_HH-MM-SS-}.  Files \texttt{clip.json} 
contain the properties \texttt{start} and \texttt{end} that indicate the real start 
and end of the video-taped sequence as marked by a local representative who watched 
the video. Files \texttt{face.zip} contain face data chunks and the JSON file with 
the video information.  Files \texttt{pose.zip} contain the OpenPose data. Zip files 
may have the extension \texttt{.gpg}, indicating that they have been encrypted with 
GNU Privacy Guard for export.

\end{document}